\documentclass[letterpaper, 10 pt, conference]{ieeeconf}

\usepackage{color}
\usepackage{soul} 
\usepackage{graphicx} 
\usepackage{multirow}
\usepackage{array}
\usepackage{tabularx}
\usepackage{hyperref}

\IEEEoverridecommandlockouts
\title{\LARGE \bf
Model Cards for AI Teammates: Comparing Human-AI Team Familiarization Methods for High-Stakes Environments \thanks{The authors thank Trent McMullen for his insightful conversations.\\All authors are with the College of Engineering at the Georgia Institute of Technology, North Avenue, Atlanta, GA, 30332, USA.\\\{bowers32, richard.agbeyibor, kolb, karen.feigh\}@gatech.edu}
}

\author{Ryan Bowers, Richard Agbeyibor, Jack Kolb, and Karen M. Feigh
}

\begin{document}
\maketitle
\thispagestyle{empty}
\pagestyle{empty}

\begin{abstract}
We compare three methods of familiarizing a human with an artificial intelligence (AI) teammate (``agent'') prior to operation in a collaborative, fast-paced intelligence, surveillance, and reconnaissance (ISR) environment. In a between-subjects user study (n=60), participants either read documentation about the agent, trained alongside the agent prior to the mission, or were given no familiarization. Results showed that the most valuable information about the agent included details of its decision-making algorithms and its relative strengths and weaknesses compared to the human. This information allowed the familiarization groups to form sophisticated team strategies more quickly than the control group. Documentation-based familiarization led to the fastest adoption of these strategies, but also biased participants towards risk-averse behavior that prevented high scores. Participants familiarized through direct interaction were able to infer much of the same information through observation, and were more willing to take risks and experiment with different control modes, but reported weaker understanding of the agent's internal processes. Significant differences were seen between individual participants' risk tolerance and methods of AI interaction, which should be considered when designing human-AI control interfaces. Based on our findings, we recommend a human-AI team familiarization method that combines AI documentation, structured in-situ training, and exploratory interaction.

\end{abstract}

\section{INTRODUCTION}
Governments have long sought to reduce reliance on human operators in high-stakes domains such as aircraft surveillance, coastal scanning, and mountainous search-and-rescue. Simultaneously, the capabilities of deep learning techniques and accessibility of high-powered compute resources have made autonomous teammates technically viable for many use cases. In recent years governments have begun supporting research to apply embodied artificial intelligence (AI) platforms to reduce the number of humans sent into high-risk scenarios by increasing the level of authority granted to AI systems in human-AI teams. 
This evolution introduces novel challenges in implementing core teamwork functions that are not typically required in traditional robotic systems. 
These functions include joint task planning, mutual performance monitoring, and adaptive behavior coordination. This challenge is magnified in high-risk domains, where teamwork principles differ significantly from those in non-expert teams \cite{klein1986rapid} and missions require high reliability and adaptability to unforeseen changes. 
The combination of increasingly autonomous systems and high-risk operational environments creates significant research challenges.

\begin{figure*}
   \includegraphics[width=\linewidth]{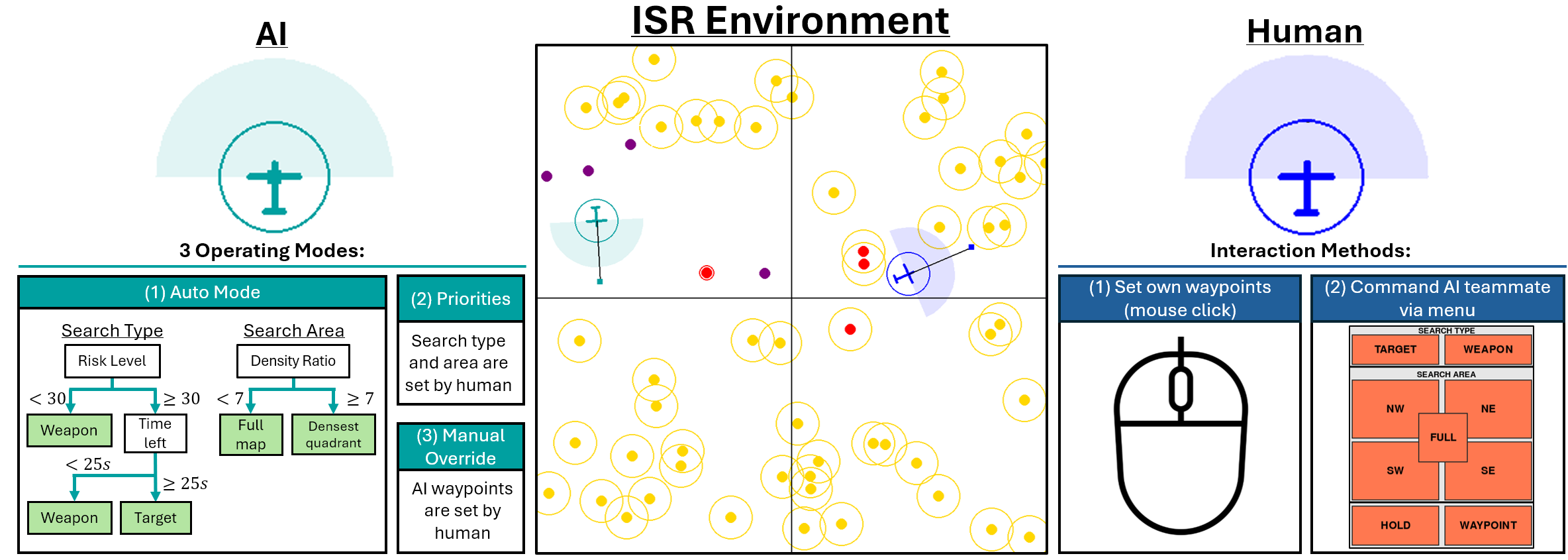}
    \caption{Overview of the ISR game environment used in the study. In this environment, a human participant and AI teammate (``agent'') collaborated to identify targets. Participants interacted with the environment by setting waypoints via mouse clicks and sending commands to the agent using a button panel. The dark blue aircraft is controlled directly by the participant. The teal aircraft is the agent, controlled by a heuristic policy. Dots represent search targets that are unknown (yellow), hostile (red), or neutral (purple). Hostile targets' weapon ranges are shown as red rings, which are yellow until they are identified.}
    \label{fig:isr}
\end{figure*}

Although recent literature has extensively mapped the essential teamwork functions required for effective human-AI collaboration \cite{feighSMM,lyonsHAT}, the adaptation of these functions to high-risk domains with expert human operators remains understudied. 
In addition, the technical implementation of these adapted functions presents significant engineering challenges, and the literature lacks study on the broader architectural elements of human-AI teaming structure in areas including authority distribution, role allocation, and team familiarization \cite{NAS2022}.

Our work focuses on one aspect of this literature gap: the process of human-AI team familiarization. 
In human-human teams, research has demonstrated a strong correlation between inter-teammate familiarity and team performance. 
This familiarity includes understanding of team members' backgrounds, capabilities, limitations, decision-making processes, and behavioral patterns. 
However, translating familiarization mechanisms to human-AI teams presents unique challenges. 
Unlike human-human teams, where team members share an inherent understanding of human cognition and behavior, human-AI teams lack this common ground. 
This suggests that an explicit understanding of AI decision-making processes may be useful for effective team performance. The need for familiarization is closely linked to explainable AI research. Whereas explainability seeks methods to explain an AI's decisions, familiarization involves presenting this information to the human to develop their mental model of the AI. 

Traditional approaches to human-AI familiarization have ranged from explicit knowledge transfer through technical documentation, to experiential learning through direct interaction \cite{Srivastava2024}. 
However, the effectiveness of these approaches in high-stakes domains remains unclear, particularly where the consequences of misunderstanding can be severe. 

This work addresses these challenges by exploring how human operators can effectively train and familiarize themselves with AI teammates in a high-stakes environment. 
Here we use \textit{high-stakes} to refer to domains in which poor task performance can lead to injury, death, or significant loss of resources.

Specifically, this work investigates the following questions:

\begin{enumerate}
    \item What knowledge about an AI teammate most effectively supports human-AI team coordination?
    \item Which methods are most effective in familiarizing a human with their AI teammate?
    \item How does human-AI teammate familiarization affect the mission strategies employed by the team?
    \item How does human-AI teammate familiarization affect team performance?\\
\end{enumerate}

Through a user study built upon a custom intelligence, surveillance, and reconnaissance (ISR) simulation environment (Fig. \ref{fig:isr}), participants were exposed to one of three familiarization methods with the agent teammate, and then completed four ISR rounds alongside that teammate. We explore the effects of the familiarization conditions on the teams' performance, resulting teamwork strategies, and user situation awareness, and identify the information about the agent that is most valuable to the human operator.

All source code and data from our study is available at \url{https://github.com/gt-cec/maisr}.

\section{Background}
\subsection{Team Training}
A team is defined as ``a distinguishable set of two or more people who interact dynamically, interdependently, and adaptively toward a common and valued goal'' \cite{salas1992toward}. In addition to individual training, team training is necessary to unleash the full capabilities of heterogeneous teams. Team training seeks to leverage the specific knowledge, skills, and attitudinal competencies of team members \cite{Salas2008}. Team training improves performance \cite{Salas2008}, decreases errors \cite{Wiener1995}, and develops team cognition \cite{Marks2002}. Research on team training has a long history, with a traditional focus on human-human teams. 

Humans perceive AI teammates as fundamentally different from human teammates \cite{Zhang2021}. Unlike human-human teams, where members share a baseline similarity in cognition and behavior, human-AI teams require deliberate strategies to overcome fundamental differences between human and AI agents. As such, research is required to determine new methods, mechanisms, and modalities of team training to realize the full capability of human-AI teams \cite{NAS2022}. These differences highlight the need for tailored training approaches to address the unique dynamics of human-AI teams.

\subsection{Training Content}
Team training involves two core elements: taskwork and teamwork. Taskwork refers to the technical and procedural knowledge required to perform specific tasks. Teamwork encompasses the skills and processes necessary for team members to effectively coordinate their actions and maintain cohesion \cite{Mohammed2010}, and can be further broken down into (1) teamwork-related processes and (2) familiarization with teammates. 

Teamwork-related processes include the operator's understanding of their own role in the team (such as mission commander or data analyst) and the roles of their teammates, standard communication protocols, shared situation awareness, and other coordination strategies. All of these elements must be incorporated into an effective team training protocol. 

Familiarization with teammates is equally vital. Operators must learn to interpret teammates' actions, understand their capabilities and limitations, and anticipate their behavioral patterns to establish effective coordination. In human-AI teams, this also includes explicitly understanding how AI agents process inputs, prioritize tasks, and adapt to dynamic environments. 

Effective training can address these challenges by incorporating documentation-based training, simulation-based exercises, and experiential learning \cite{GormanCooke2010}. Such approaches provide operators with the experience needed to adapt their strategies to the unique dynamics of human-AI collaboration.

\subsection{Team Familiarization / Habituation}
\textit{Familiarization}, or habituation, refers to the process through which team members gain a deeper understanding of their teammates' capabilities, decision-making processes, and operational behaviors. In human-human teams, even heterogeneous teams where each team member has a specialized role, members share an inherent understanding of each others' behavior and mental processes. High-performing teams, especially in high-risk domains, tend to have a strongly aligned understanding of team members' backgrounds, capabilities, and behavioral patterns \cite{Salas1995}. 

In human-AI teams, teamwork becomes more complex due to inherent differences between human and AI cognition \cite{christensengraybox}, and the absence of shared cognitive frameworks and natural communication channels. Unlike human-human teams, where familiarization and shared cognitive frameworks often emerge naturally \cite{cannon-bowers1993shared}, familiarization in human-AI teams requires deliberate effort through training, interaction, and system transparency \cite{Srivastava2024}.

Human-AI team training must explicitly address these gaps by equipping human operators with strategies to interpret AI behavior and coordinate effectively \cite{NAS2022}. This includes (1) developing an accurate mental model of the AI teammate's capabilities, limitations, and behavior, and (2) adapting teamwork processes to account for the AI's decision-making mechanisms and communication methods.

\section{System Design}

\subsection{Multi-Agent ISR Game Environment}

We introduce a 2D Pygame environment that simulates a cooperative intelligence, surveillance, and reconnaissance (ISR) task. In this environment, a team of two aircraft (one controlled by a human, one flown by a rule-based AI policy) must explore the map to identify targets as either neutral or hostile. Hostile targets have weapons that must also be identified by flying closer to the target. Both aircraft start with 10 health points (HP) and take damage if they fly inside a weapon's range. The team's goal is to identify as many targets and weapons as possible within the four-minute time limit without losing all health points. This environment is extended from prior work \cite{agbeyibor2024towards, agbeyibor2024joint}.

Each aircraft has two sensors: a long-range sensor for identifying targets, and a short-range sensor for identifying weapons. The aircraft sensor ranges and hostile weapon ranges are balanced such that target identification can be completed without taking damage, but weapon identification requires the aircraft to enter the weapon's range and potentially take damage. This creates an asymmetric risk profile between the two search tasks. 

\subsection{The AI Teammate}
The AI teammate (``agent'') follows a heuristic path-planning policy that computes a desired flight path using waypoints (\textit{x,y} location), which the aircraft navigates towards using a low-level controller. By default, the agent flies towards the nearest \textit{valid} target, where target validity is determined by two types of search priorities:

\begin{itemize}
    \item \textbf{Search Type}: The agent may search for \texttt{targets only}, or \texttt{targets and weapons}.
    \item \textbf{Search Area}: The agent may prioritize a specific map quadrant (\texttt{NW, NE, SW, SE}), or the \texttt{full map}.
\end{itemize}

The agent has three operating modes:

\begin{itemize}
    \item \textbf{Auto}: The agent determines its own search priorities based on two periodically-updated heuristics (Fig. \ref{fig:isr}), then autonomously searches for targets according to those priorities. \textit{Search Type} is determined by a \texttt{Risk Level} heuristic that is a function of the agent's current HP and the number of nearby hostile targets. \textit{Search Area} is determined by a \texttt{Density Ratio} heuristic that prioritizes the quadrant with the greatest number of unknown targets if it contains at least seven more targets than the next-densest quadrant.
    
    \item \textbf{Priorities}: The agent's \textit{Search Type} and \textit{Search Area} are manually set by the human, and the agent autonomously searches for targets according to those priorities. 

    \item \textbf{Override}: The human can command the agent to fly towards a specific waypoint or hold in place. While under manual control the agent's normal path planning algorithms are disabled. 
\end{itemize}

In a real-world scenario, the unmanned aircraft would likely have a different form factor from a traditional crewed aircraft (typically smaller). To reflect this, we configured the agent's aircraft to be more resilient to hostile weapon damage than the human's aircraft. The design has a secondary advantage of introducing a unit-level optimization for the team's strategies that participants can leverage.

\subsection{Human-AI Interaction Methods}
The ISR environment allows the human to send commands to the agent to modify its behavior at any time during a round. Any command may be given at any time based on the human's desired strategy. In this way, the human has two roles -- performing the primary task of searching for targets, and coordinating the team's strategy using commands. The available commands are:

\begin{itemize}
    \item \textbf{Search Priorities:} Specify the agent's \textit{Search Type} and \textit{Search Area} priorities. Causes the agent to enter \texttt{Priorities} mode. 
    \item \textbf{Hold}: Stop in the current location. Causes the agent to enter \texttt{Override} mode.
    \item \textbf{Waypoint}: Fly to a specific waypoint. Causes the agent to enter \texttt{Override} mode.
    \item \textbf{Auto}: Set the agent to \texttt{Auto} mode. 
\end{itemize}

\subsection{Team Strategy Elements}

This environment lends itself to a variety of strategies, which we define along three axes:

\begin{enumerate}

    \item \textbf{Control Mode}: The human may control the agent at a high (\texttt{Auto} mode), medium (\texttt{Priorities} mode), or low (\texttt{Override} mode) level. They may also use a combination of the three modes, moving between modes as desired.
    
    \item \textbf{Risk Distribution}: Teams may distribute risk symmetrically by equally sharing the weapon identification task (the main source of mission risk), or asymmetrically by focusing one member on target identification and one member on weapon identification. Since the agent's aircraft is less vulnerable to weapon damage, the human has an incentive to prioritize target identification and command the agent to focus on weapon identification.

    \item \textbf{Spatial Coordination}: Teams can use a variety of spatial coordination methods, such as dividing the map to cover as much ground as possible, or searching the map together in close formation.
\end{enumerate}

Notably, there is no obvious \textit{best} strategy for the environment. We anticipate that players will converge upon a strategy that best fits their understanding of the agent and maximizes the team's performance objective.



\section{METHODS}
We conducted a between-subjects user study (n=60, 25\% female, ages 18-39) in which participants worked alongside the agent to complete four rounds in the ISR environment.

\subsection{Participants}
We recruited participants from a local university, including both students and faculty. Inclusion criteria included a minimum of one hour per week of current or past video game activity. Among the 60 participants recruited, 55 reported more than 5 hours per week of current or previous video game activity. Participants were compensated for their time and the top 20\% received a 25\% bonus.

Prior to attending the study, participants complete a demographics questionnaire regarding their video game experience (hours/week activity and type of video games played), age, gender, prior AI and machine learning (AI/ML) experience, and trust in automation (assessed using the adapted Propensity to Trust Technology scale from \cite{jessuppropensity}). The questionnaire results were used to balance participants across the study conditions.

\subsection{Experiment Design}
Participants were placed in one of three condition groups. Each group received a different type of familiarization with the agent before beginning the four ISR rounds:

\begin{itemize}
    \item \textbf{Documentation}: Participants received a 6-page document that described the agent's path planning, strengths and weaknesses, decision-making algorithms, and technical details (e.g. observation and action space). This document is inspired by the model cards commonly used to describe machine learning models \cite{mitchell2019model}.
    
    \item \textbf{In-situ}: Participants completed a training round with the agent before the main four rounds (the other two groups' training round did not include the agent). The training round was a no-risk opportunity to observe the agent and experiment with different commands.
    
    \item \textbf{Control}: The Control group received the least familiarization with the agent. Participants were told that the agent would help search for targets, and were shown how to command the agent, but they did not receive detailed technical or behavioral information.
\end{itemize}

The Documentation and In-situ groups will be referred to collectively as the Familiarization groups.

\subsection{Dependent Variables}

\begin{figure}
    \centering
    \includegraphics[width=0.6\linewidth]{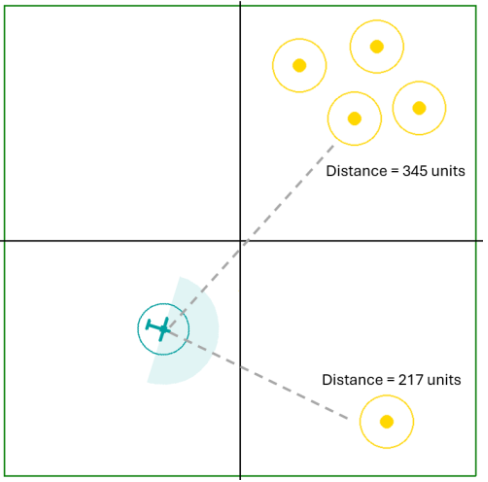}
    \caption{A vignette from the AI Understanding quiz. For this question, participants could select ``The agent will fly toward the single target in the bottom right'' or ``the agent will fly toward the cluster of targets.''}
    \label{fig:vignette-q1}
\end{figure}

To evaluate team performance and strategies we obtained a variety of metrics:

\begin{enumerate}
    \item \textbf{Mission Performance}: Performance was tracked using a point system. The team earned points for each identified target (+10), each identified weapon (+5), completing a round with health remaining (+70 per HP), and completing a round with time remaining (+15 per second). Points were deducted if the human (-400) or agent (-300) were destroyed.

    \item \textbf{Strategy Metrics}: The participant's strategy was monitored using the percentage of time the agent was in each command mode, the percentage of targets and weapons identified by the human compared to the agent, and the percentage of time that the human and the agent were located within the same quadrant.

    \item \textbf{Situation Awareness}: We used the Situation Awareness Global Assessment Technique (SAGAT) \cite{endlseySagat} to assess participant situation awareness (SA) at each minute during the game rounds. SA questions included the current location of the human and agent on the map, and each player's current health level. Each round included three SAGAT surveys, for a total of 12 surveys and 48 questions per participant.

    \item \textbf{Workload}: Participants completed a NASA TLX \cite{hart1988development} questionnaire after the last round.

    \item \textbf{Cognitive Trust in the Agent}: Participants completed a custom 17-23 question Likert scale ($\alpha=0.932$) after the fourth round to assess their trust in the agent's competence and reliability in the ISR mission. 

    \item \textbf{AI Understanding Quiz}: We assessed participants' understanding of the agent's behavior using a 3-question quiz. Each question asked what the agent would do in a given scenario. An example is shown in Fig. \ref{fig:vignette-q1} and other questions are listed in Table \ref{tab:quiz-questions}.
\end{enumerate}


\begin{table}[htbp]
\vspace{+7pt}
\centering
\caption{AI Understanding Quiz Questions}
\label{tab:quiz-questions}
\vspace{-5pt}
\begin{tabular}{lcp{5cm}}
\hline
\# & Topic & Details \\
\hline
1 & Target priorities & Will the agent prioritize a distant cluster of targets or a single nearby target? \\
2 & Path planning & Will the agent detour to identify nearby targets while enroute to a commanded quadrant? \\
3 & Area priorities & Will the agent prioritize a quadrant that has 6 more targets than the others? \\
\hline
\end{tabular}

\end{table}


\subsection{Hypotheses}

We hypothesize that, compared to the Control group, participants in the familiarization groups will: 

\begin{itemize}

    \item \textbf{H1}: Achieve higher overall ISR game scores.
    
    \item \textbf{H2}: Be more likely to command the agent to identify weapons (based on knowledge of the agent's relative strengths and weaknesses).
    
    \item \textbf{H3}: Have a more accurate mental model of the agent's behavior, as measured by their score on questions about the agent's behavior in specific scenarios. 
    
    \item \textbf{H4}: Have a lower workload from performing the task.
\end{itemize}

\subsection{Experiment Procedure}

The experiment began with a briefing that described the ISR game mechanics, objectives, the human’s role, the agent, and the commands the human can give the agent. After the briefing, the Documentation group read the agent's model card.

Next, participants completed a training round. The training round included the agent for the In-situ group, and only the human aircraft for the other groups.

All participants then completed four 4-minute ISR game rounds that were periodically paused to administer the SAGAT survey. All participants saw the same four map layouts. After the fourth round, participants completed a final survey to assess their trust in the agent, their overall workload (NASA TLX), and their understanding of the agent.

After the final survey, we interviewed participants about their experience with the agent and the strategies they used in the game. The interview also included targeted questions to identify what information about the agent was helpful. The interview directly referenced the model card if the participant was in the Documentation group, and referenced what the participant learned about the agent through observation if they were in the Control or In-situ groups.

\section{RESULTS}

\subsection{Value of Information about the Agent}

Qualitative analysis of the interview comments revealed that participants primarily found two categories of information about the agent valuable. First, understanding the agent's decision-making processes, particularly its path planning and target prioritization, allowed participants to anticipate the agent's behavior and use commands to align it with their desired strategies. This was noted by 23 participants (7 Documentation, 16 other conditions).

Second, knowledge of the agent's strengths and weaknesses (specifically, that it could avoid weapon damage) allowed participants to allocate team tasks effectively. This resulted in participants commanding the agent to identify weapons and providing quadrant and waypoint commands to guide the agent along a more efficient global search route. This was noted by 13 participants (9 Documentation, 4 other conditions).

Technical details like the agent's rule-based nature and observation space were also valuable to participants with AI/ML experience.

Some information was decidedly not helpful -- most participants deemed the agent's two priority heuristic formulas to be too complex to be useful during gameplay.


Each group applied this knowledge about the agent differently. The Documentation group developed a more accurate understanding of the agent's design compared to other groups. When asked to describe how the agent worked, 8 Documentation participants (versus 4 in other groups) referenced the agent's internal processes multiple times as the primary source of its actions. However, 8 Documentation participants also reported needing 1-2 rounds before they could operationalize the model card's information.

In contrast, the In-situ group discussed the agent in behaviorist terms (15 In-situ, 11 Control, 8 Documentation), with less understanding of its internal state. However, they reported greater understanding of the agent's Auto mode, with 8 In-situ participants reporting that Auto mode was effective (compared to 2 in other groups). Many In-situ participants used the training round to experiment with Auto mode, and therefore were more willing to use this mode in the four main rounds. Conversely, 6 participants in the Documentation and Control group reported being hesitant to use Auto mode because they did not understand it and were not willing to risk experimenting with it in the main rounds.

The Control group reported the least understanding of the agent. 4 Control participants (compared to 1 Documentation and 1 In-situ) reported viewing the agent as black box and discussed its role in generic terms that did not consider the specifics of its behavior.




\subsection{Differences in Team Strategies}

There were statistically significant differences in the game strategies used by the three condition groups. The differences were most significant in the first two rounds, as the groups converged towards similar strategies in the final two rounds. 

Team strategies were assessed using a standard 1 way, 3 level ANOVA where the assumptions for normality held. Where normality assumptions were violated, a Kruskal-Wallis non-parametric test was substituted. A paired t-test was used to compare differences between rounds within a group.  Across all tests, significance was set at the 0.05 Type I error level.

\subsubsection{Control Modes} 
\begin{figure}
   \includegraphics[width=\linewidth]{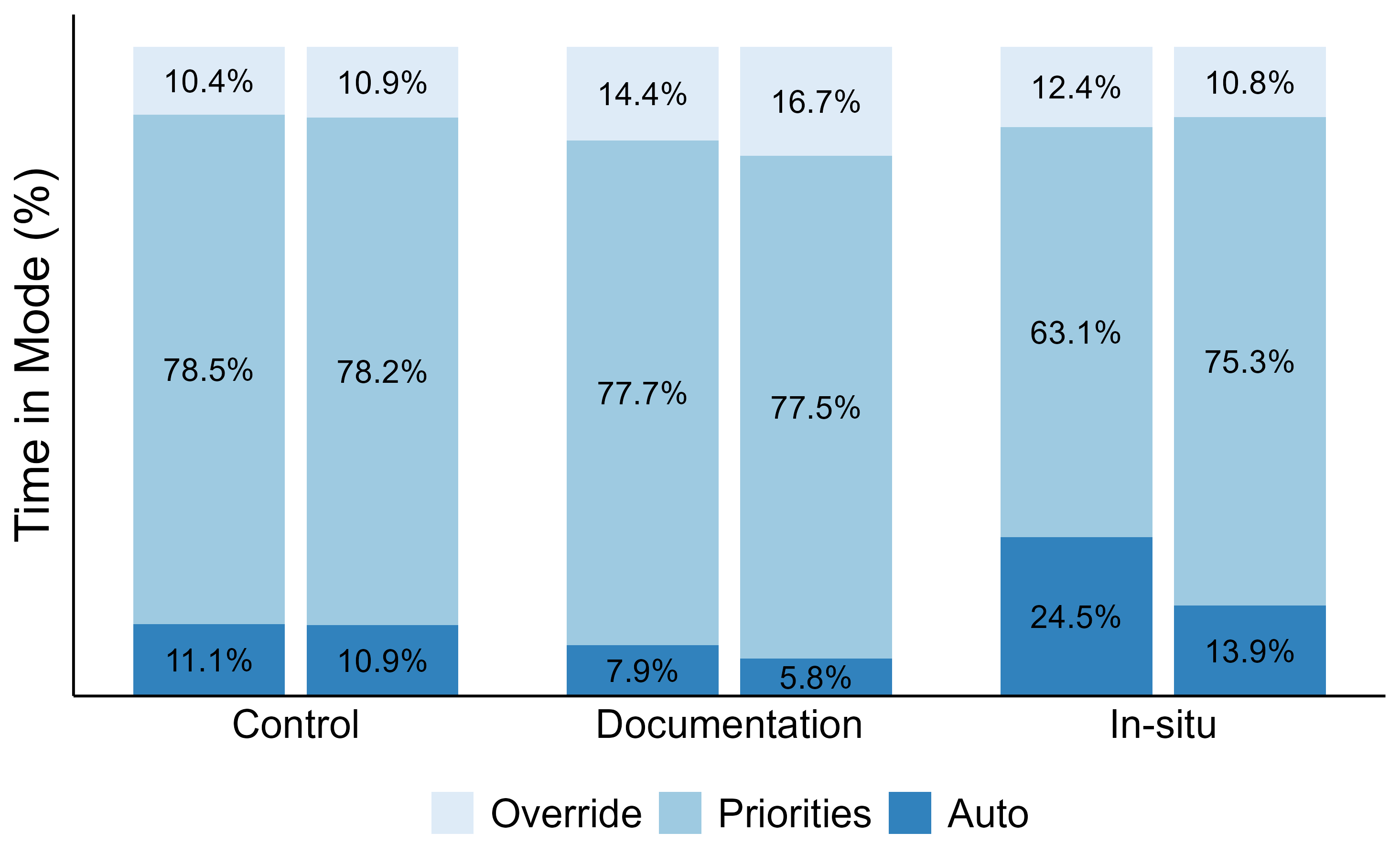}
   \vspace{-20pt}
    \caption{Breakdown of time spent using each control mode by each group, split between rounds 1-2 (left) and rounds 3-4 (right). }
    \label{fig:level-of-control-earlylate}
\end{figure}

All three groups overwhelmingly preferred Priorities mode, covering $>70\%$ of all round time, as shown in Fig. \ref{fig:level-of-control-earlylate}. 

The Documentation group exerted the most direct control over the agent, using Override mode more (Kruskal-Wallis, $\chi^2=6.94$ $p=0.031$) and Auto mode less (ANOVA, $F(1,3)=3.97$ $p = 0.025$) than other groups. The In-situ group was more likely to use Auto mode, especially in the first two rounds (ANOVA $F(1,3)=4.59$, $p=0.014$).

In later rounds, both familiarization groups increased their use of Override and Priorities mode (and decreased use of Auto mode). This change was most significant in the In-situ group (Paired t-test, $t=2.1$, $p = 0.05$), while the Control group showed no change.





\subsubsection{Risk Distribution} 
\begin{figure}
   \includegraphics[width=\linewidth]{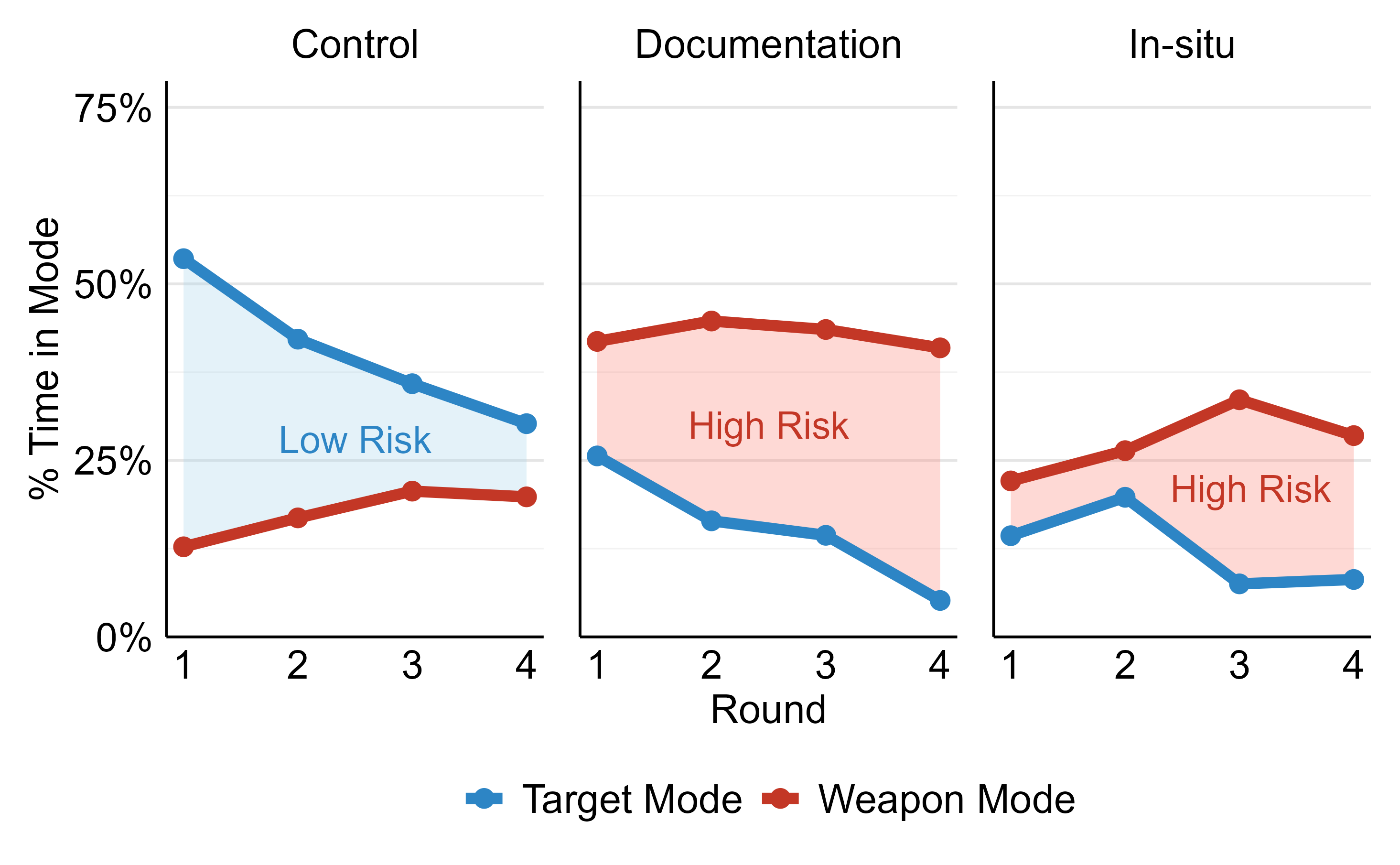}
   \vspace{-25pt}
    \caption{Comparison of each group's propensity to command the agent to identify targets (Target mode) vs. to identify weapons (Weapon mode). Lines are drawn between each group's mean value in each round.}
    \label{fig:weaponmode-evolution}
\end{figure}

\begin{figure}
   \includegraphics[width=\linewidth]{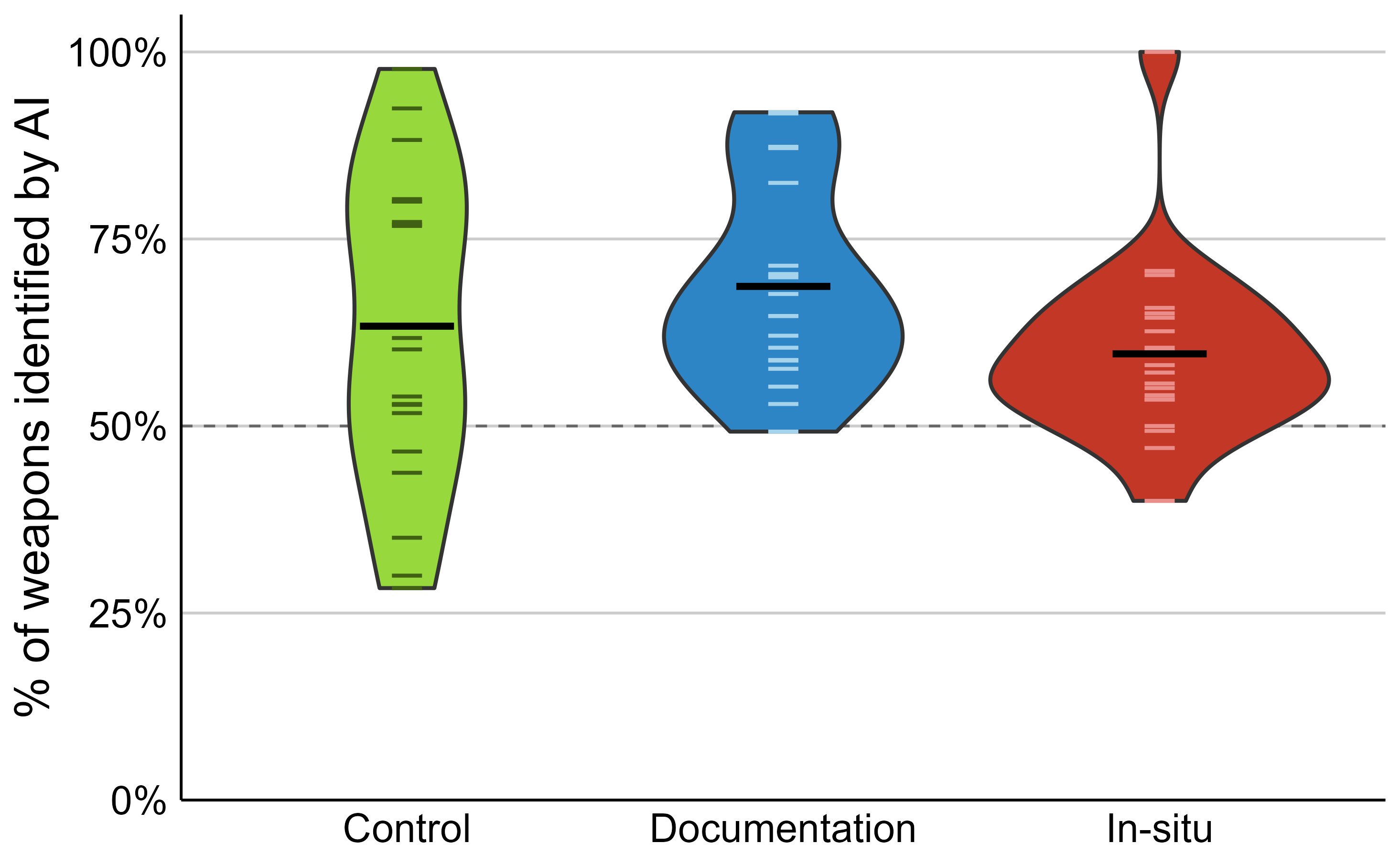}
   \vspace{-20pt}
    \caption{The percentage of team weapon IDs that were performed by the agent. Higher percentages suggest that the participant had a higher tolerance for risk to the agent than to themselves.}
    \label{fig:ai-weaponid-percentage}
\end{figure}

Both familiarization groups (Documentation and In-situ) were more likely to command the agent to identify weapons (Wilcoxon, $p=0.021$) and less likely to command the agent to identify targets (ANOVA, $p=0.0007$) than the Control group, as seen in Fig. \ref{fig:weaponmode-evolution}. The Documentation group also allowed the agent to identify a greater share of weapons (Kruskal-Wallis, $p=0.115$), as seen in Fig. \ref{fig:ai-weaponid-percentage}. These differences are more significant in the first two rounds (Kruskal-Wallis, $p=0.0199$). Thus, we affirm hypothesis \textbf{H2}.


\subsubsection{Spatial Coordination} Spatial coordination strategies were similar between groups. Most participants preferred to split the map in half, commanding the agent to search one half of the map while the human searched the other. The alternative (less common) strategy was for the human and agent to search the map as a team, staying close together for most of the round. This ``tag team" tactic tended to result in low scores, while the ``divide-and-conquer" strategy included high and low scorers depending on other elements of the participant's strategy.



\subsection{Performance Differences Between Groups}

Average ISR round scores were similar between the three groups (Table \ref{tab:group-game-performance}). The In-situ and Control groups showed a larger score variance than the Documentation group, but this difference was not statistically significant (Levene's test, $p = 0.184$). Thus, we refute hypothesis \textbf{H1}.

\begin{table}[htbp]
\centering
\caption{Mean and Standard Deviation of Round Scores}
\label{tab:group-game-performance}
\vspace{-5pt}
\begin{tabular}{lccccc}
\hline
Group & Average Score & Standard Deviation \\
\hline
Control & 1275 & 236 \\
Document & 1287 & 140 \\
In-situ & 1295 & 253 \\
\hline
\end{tabular}

\end{table}



\begin{figure}
    \centering
    \includegraphics[width=\linewidth]{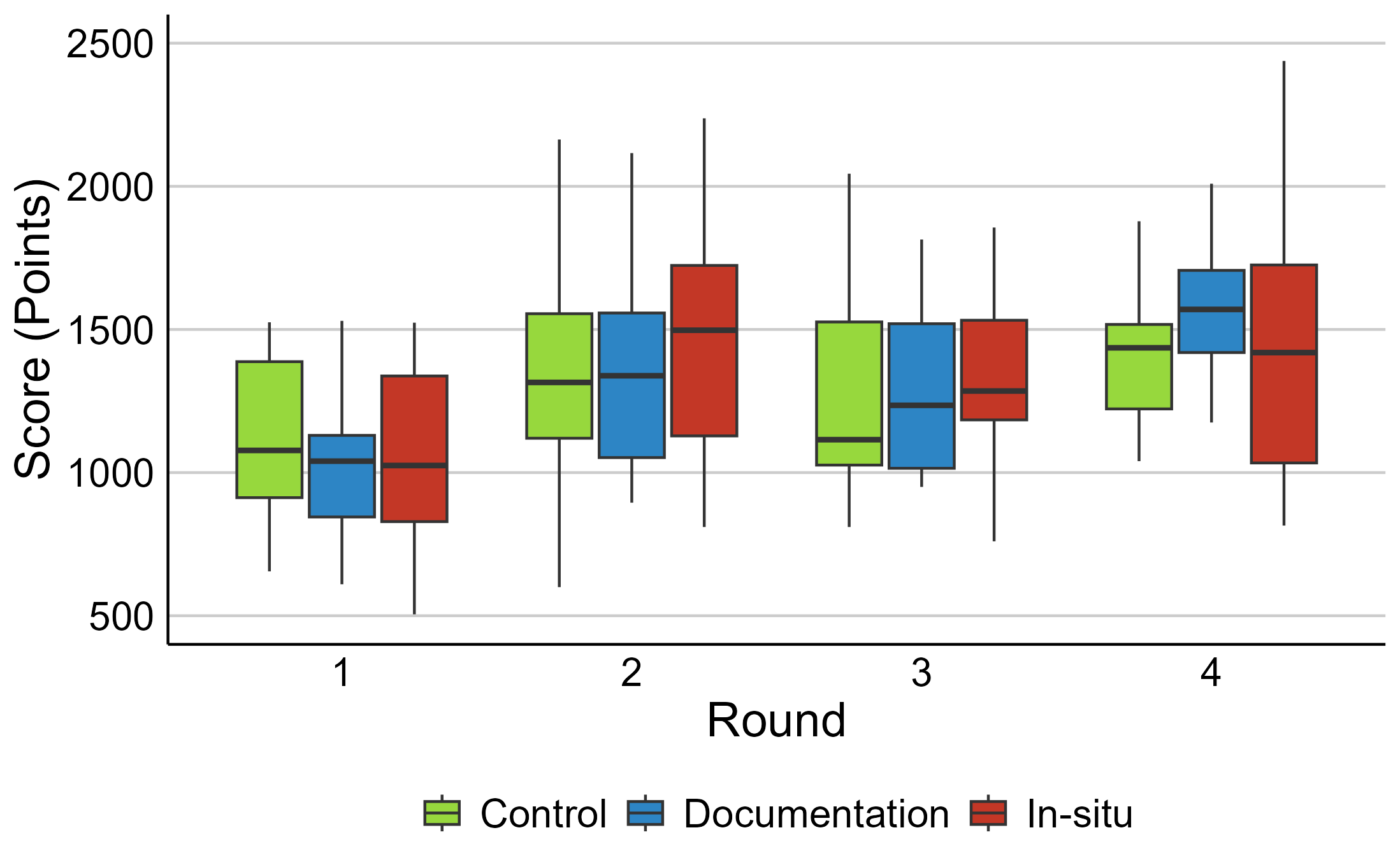}
    \vspace{-25pt}
    \caption{Group scores from round 1 to 4. Compared to the Control and In-situ groups, the Documentation group's score decreased less from round 2 to 3 and they performed better in round 4.}
    \label{fig:score-growth}
\end{figure}

The Documentation group's score consistency is also reflected in the growth in scores across the four rounds.  As shown in Fig. \ref{fig:score-growth}, the Documentation group's score increased more consistently than the other groups, achieving the highest scores over all groups in the final round.

\subsection{AI Understanding Quiz Scores}

The Familiarization groups outperformed the Control group on the AI Understanding quiz, as shown in Table \ref{tab:group-quiz-performance-1}, but this was not statistically significant (Kruskal-Wallis, $p=0.37$), so we cannot reject the null hypothesis for \textbf{H3}. Individual question scores indicate that In-situ participants correctly inferred the agent's path planning (even outperforming the Documentation group), but typically did not understand its target selection algorithm.

\begin{table}[htbp]
\centering
\caption{AI Understanding Quiz Scores by Group}
\label{tab:group-quiz-performance-1}
\vspace{-5pt}
\begin{tabular}{lccccc}
\hline
Group & Quiz & \% Correct & \% Correct & \% Correct \\
 & Score & Q1 & Q2 & Q3 \\
\hline
Control & 1.50 & 45.0 & 75.0 & 30.0 \\
Document & 1.85 & 60.0 & 80.0 & 45.0 \\
In-situ & 1.84 & 47.4 & 89.5 & 47.4 \\
\hline
\end{tabular}
\end{table}

\subsection{Situation Awareness and Workload}
\begin{figure}
   \includegraphics[width=\linewidth]{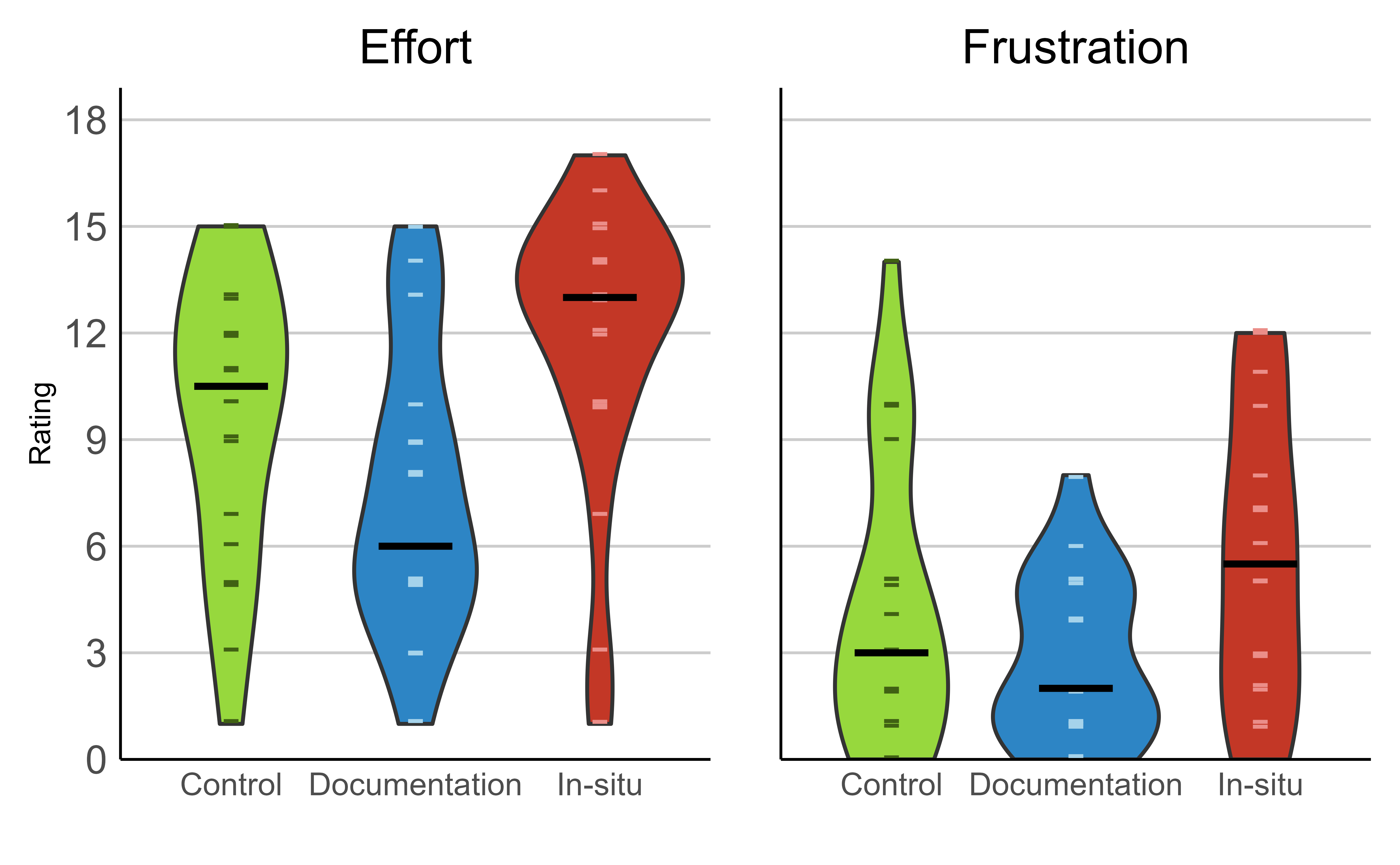}
   \vspace{-20pt}
    \caption{NASA TLX frustration and effort ratings between groups.}
    \label{fig:workload-effort-frustration}
    \vspace{-10pt}
\end{figure}

Situation Awareness scores were similar between groups. The In-situ group reported slightly higher composite workload scores (ANOVA, $p=0.12$), and significantly higher effort ($p=0.016$) and frustration ($p=0.027$) than other groups, as seen in Fig. \ref{fig:workload-effort-frustration}. Thus we refute hypothesis \textbf{H4}. Increased frustration was also positively correlated with the human identifying more weapons than the agent, finishing a round early, and the amount of damage taken by both players during the round.

\subsection{Effects of AI/ML Experience} \label{ai-experience}

Participants with AI/ML experience were less likely to command the agent to identify weapons (25\% vs. 35\% of the time, ANOVA, $p=0.192$). Within the familiarization groups, participants with AI/ML experience performed worse on the AI Understanding quiz (1.45 vs. 2.11), while in the Control group, those with AI/ML experience outperformed those without (1.69 vs. 1.14, Wilcoxon, $p = 0.11$).



\section{Discussion}

\subsection{Familiarization-induced Emergent Strategies}

Familiarization increased participant tolerance for risk to the agent, as they were more likely to command the agent to identify weapons. It is noteworthy that the Control group also began to adopt this strategy in later rounds (Fig \ref{fig:weaponmode-evolution}). This indicates that familiarization helped participants discover and habituate to this ``convergent strategy" more rapidly. The Documentation group took this even further: these participants tended to focus on identifying targets while assigning weapon identification entirely to the agent, thereby delegating all mission risk to the agent.

This risk delegation explains why the Documentation group had fewer score outliers compared to other groups. The two most impactful score components are the human's HP remaining (up to a +700pt bonus) and time remaining (the maximum observed was +1260pts for finishing 84 seconds early) at the end of a round. The time bonus is only earned if all targets and weapons are identified, and it is not possible for the agent to identify all weapons within the time limit if acting alone. Thus, if the participant hopes to earn the time bonus, they must also take some risk onto themselves to identify weapons. This risk often did not pay off -- many participants lost more points from damage than they earned for finishing early -- however, top participants were able to earn both bonuses.

Meanwhile, participants who delegated all risk to the agent typically scored around 1300 points (600pts for identifying all targets plus 700pts for full HP remaining). Most of these low-risk players were in the Documentation group, which explains the lack of score outliers in that group. Thus, while documentation-based familiarization caused the fastest adoption of the convergent strategy, it also led to an over-fixation on this strategy, which prevented many participants from reaching higher scores.

The In-situ group was more likely to experiment with different command modes, evident by their higher use of Auto mode compared to other groups. Interview comments confirmed that participants avoided Auto mode if they did not understand it, so the training round proved to be a valuable opportunity for the In-situ group to learn about the agent's behavior. This suggests that exploratory interaction may be a crucial element of human-AI familiarization, allowing the human to probe the agent's behavior and develop their own interaction style in a low-threat setting.

\subsection{Preferred Level of Control over the Agent} 
We found a substantial range of preferences for level of control over the agent. Some participants controlled the agent directly using waypoints, while other participants did not send the agent a single command across all four rounds. Level of control did not significantly impact the team's score.

Notably, seven participants strongly preferred a counterproductive strategy of micromanaging the agent using excessive waypoint commands, to the extent of neglecting their own aircraft. These participants often became overwhelmed, took substantial damage to their own aircraft, or even failed the round. While the finding indicates that the environment is sufficiently complex to discourage micromanaging, we are interested in the root cause of this micromanagement. Future work may explore whether a correlation exists with certain personality traits, and whether these traits are common in people who choose careers in high-stakes domains.

\subsection{Misconceptions about the Agent}
We had anticipated that prior AI/ML experience would positively correlate with performance on the AI Understanding quiz. However, we instead found a weak but opposite trend. This may be due to preconceptions (informed by prior AI/ML experience) about heuristic policy behavior, which did not always apply to the agent. Meanwhile, AI/ML \textit{non}-experts voiced common AI/ML misconceptions, such as assuming all ML algorithms involve online learning.



\section{Conclusions}
In this work we compared 
three methods of familiarizing a human operator with an autonomous AI teammate. We conducted a user study with $20$ participants in each condition, and found that the different familiarization methods had meaningful impacts on participant strategy and risk delegation.

We found that documentation-based and experience-based familiarization with the agent allowed participants to quickly habituate to sophisticated team strategies that leveraged each member's unique strengths and weaknesses. Documentation-based familiarization provided a strong theoretical foundation that led to the fastest strategy habituation, but biased participants toward overly conservative strategies and required several rounds of gameplay for participants to fully internalize the information. In-situ familiarization allowed participants to experiment with diverse strategies that often led to high scores, but a less developed mental model of the agent. We recommend a hybrid process that combines documentation, structured in-situ training, and exploratory interaction. 

In post-hoc interviews, many participants stated that they wanted to understand their agent's decision-making processes (to predict its behavior) and its strengths and weaknesses (to determine an optimal task allocation). This mirrors human-human teams -- teams often specialize along what each member is good at, and it is common for team members to deliberately seek out this information as they become familiarized with each other.
    
Participants with AI/ML experience desired more technical information about the agent than those without AI/ML experience. Future work may explore adapting the technicality of the information to the subject's level of expertise. In addition, participants without AI/ML experience held common misconceptions about AI, while AI/ML experts often incorrectly applied their preconceptions to the agent. System designers should address both types of misconceptions during the training process.

Lastly, individual participants varied significantly in their risk tolerance and preferred level of control over the agent. Designers of control interfaces for human-AI teams in complex environments should account for this variability and the impact of familiarization techniques on risk tolerance, particularly if tasked to develop a ``one size fits all" architecture. We anticipate that this work's insights will inform the development of more effective human-AI team training protocols for high-stakes operational environments.

\bibliographystyle{plain}
\bibliography{references.bib}

\begin{thebibliography}{10}

\bibitem{agbeyibor2024towards}
Richard Agbeyibor, Vedant Ruia, Jack Kolb, Carmen~Jimenez Cortes, Samuel Coogan, and Karen~M Feigh.
\newblock Towards safe collaboration between autonomous pilots and human crews for intelligence, surveillance, and reconnaissance.
\newblock In {\em 2024 AIAA DATC/IEEE 43rd Digital Avionics Systems Conference (DASC)}, pages 1--8. IEEE, 2024.

\bibitem{agbeyibor2024joint}
Richard Agbeyibor, Vedant Ruia, Jack Kolb, and Karen~M Feigh.
\newblock Joint intelligence, surveillance, and reconnaissance mission collaboration with autonomous pilots.
\newblock In {\em Proceedings of the Human Factors and Ergonomics Society Annual Meeting}, volume~68, pages 409--415. SAGE Publications Sage CA: Los Angeles, CA, 2024.

\bibitem{cannon-bowers1993shared}
Janis~A. Cannon-Bowers, Eduardo Salas, and Sharolyn Converse.
\newblock Shared mental models in expert team decision making.
\newblock In N.~John Castellan, editor, {\em Individual and Group Decision Making}, page~26. Psychology Press, 1 edition, 1993.

\bibitem{christensengraybox}
James~C. Christensen and Joseph~B. Lyons.
\newblock Trust between humans and learning machines: Developing the gray box.
\newblock {\em Mechanical Engineering}, 139(06):S9--S13, 06 2017.

\bibitem{endlseySagat}
M.R. Endsley.
\newblock Situation awareness global assessment technique (sagat).
\newblock In {\em Proceedings of the IEEE 1988 National Aerospace and Electronics Conference}, pages 789--795 vol.3, 1988.

\bibitem{GormanCooke2010}
Jamie~C. Gorman, Nancy~J. Cooke, and Polemnia~G. Amazeen.
\newblock Training {Adaptive} {Teams}.
\newblock {\em Human Factors}, 52(2):295--307, April 2010.

\bibitem{hart1988development}
Sandra~G Hart and Lowell~E Staveland.
\newblock Development of nasa-tlx (task load index): Results of empirical and theoretical research.
\newblock In Peter~A Hancock and Najmedin Meshkati, editors, {\em Human Mental Workload}, pages 139--183. North-Holland, Amsterdam, 1988.

\bibitem{jessuppropensity}
Sarah Jessup, Tamera Schneider, Gene Alarcon, Tyler Ryan, and August Capiola.
\newblock {\em The Measurement of the Propensity to Trust Automation}, pages 476--489.
\newblock 06 2019.

\bibitem{klein1986rapid}
Gary~A Klein, Robert Calderwood, and Anne Clinton-Cirocco.
\newblock Rapid decision making on the fire ground.
\newblock {\em Proceedings of the Human Factors Society Annual Meeting}, 30(6):576--580, 1986.

\bibitem{lyonsHAT}
Joseph~B Lyons, Katia Sycara, Michael Lewis, and August Capiola.
\newblock Human-autonomy teaming: Definitions, debates, and directions.
\newblock {\em Frontiers in Psychology}, 12:589585, May 2021.

\bibitem{Marks2002}
Michelle~A. Marks, Mark~J. Sabella, C.~Shawn Burke, and Stephen~J. Zaccaro.
\newblock The impact of cross-training on team effectiveness.
\newblock {\em Journal of Applied Psychology}, 87(1):3--13, 2002.

\bibitem{mitchell2019model}
Margaret Mitchell, Simone Wu, Andrew Zaldivar, Parker Barnes, Lucy Vasserman, Ben Hutchinson, Elena Spitzer, Inioluwa~Deborah Raji, and Timnit Gebru.
\newblock Model cards for model reporting.
\newblock In {\em FAT* '19: Conference on Fairness, Accountability, and Transparency}, pages 220--229, Atlanta, GA, USA, 2019. ACM.

\bibitem{Mohammed2010}
Susan Mohammed, Lori Ferzandi, and Katherine Hamilton.
\newblock Metaphor {No} {More}: {A} 15-{Year} {Review} of the {Team} {Mental} {Model} {Construct}.
\newblock {\em Journal of Management}, 36(4):876--910, July 2010.

\bibitem{NAS2022}
The National~Academies of~Sciences.
\newblock {\em Human-{AI} {Teaming}: {State}-of-the-{Art} and {Research} {Needs}}.
\newblock The National Academies of Sciences, Engineering, December 2021.

\bibitem{feighSMM}
Divya~Srivastava Robert W.~Andrews, J. Mason~Lilly and Karen~M. Feigh.
\newblock The role of shared mental models in human-ai teams: a theoretical review.
\newblock {\em Theoretical Issues in Ergonomics Science}, 24(2):129--175, 2023.

\bibitem{salas1992toward}
E.~Salas, T.~L. Dickinson, S.~A. Converse, and S.~I. Tannenbaum.
\newblock Toward an understanding of team performance and training.
\newblock In R.~W. Swezey and E.~Salas, editors, {\em Teams: Their training and performance}, pages 3--29. 1992.

\bibitem{Salas1995}
Eduardo Salas, Clint~A. Bowers, and Janis~A. Cannon-Bowers.
\newblock Military {Team} {Research}: 10 {Years} of {Progress}.
\newblock {\em Military Psychology}, 7(2):55--75, June 1995.

\bibitem{Salas2008}
Eduardo Salas, Deborah DiazGranados, Cameron Klein, C.~Shawn Burke, Kevin~C. Stagl, Gerald~F. Goodwin, and Stanley~M. Halpin.
\newblock Does {Team} {Training} {Improve} {Team} {Performance}? {A} {Meta}-{Analysis}.
\newblock {\em Human Factors}, 50(6):903--933, December 2008.

\bibitem{Srivastava2024}
Divya Srivastava, J.~Mason Lilly, and Karen~M. Feigh.
\newblock Exploring the role of judgement and shared situation awareness when working with {AI} recommender systems.
\newblock {\em Cognition, Technology \& Work}, July 2024.

\bibitem{Wiener1995}
Earl~L. Wiener, Barbara~G. Kanki, and Robert~L. Helmreich.
\newblock {\em Cockpit {Resource} {Management}}.
\newblock Gulf Professional Publishing, November 1995.

\bibitem{Zhang2021}
Rui Zhang, Nathan~J. McNeese, Guo Freeman, and Geoff Musick.
\newblock "{An} {Ideal} {Human}": {Expectations} of {AI} {Teammates} in {Human}-{AI} {Teaming}.
\newblock {\em Proceedings of the ACM on Human-Computer Interaction}, 4(CSCW3):246:1--246:25, January 2021.

\end{thebibliography}

\end{document}